\documentclass[runningheads]{llncs}

\usepackage{eccv}

\usepackage{eccvabbrv}

\usepackage{graphicx}
\usepackage{booktabs}
\usepackage{amsmath,amsfonts,bm}
\usepackage{subcaption}
\usepackage{wrapfig}
\usepackage{multicol}
\usepackage{multirow}
\usepackage{pifont}
\usepackage{algorithm}
\usepackage{algpseudocode}
\usepackage{array}
\usepackage{adjustbox}
\usepackage{makecell}
\usepackage{nicefrac}
\usepackage{microtype}
\usepackage{xcolor}
\usepackage{fvextra}
\usepackage[edges]{forest}
\usepackage[utf8]{inputenc}
\usepackage[T1]{fontenc}

\usepackage[accsupp]{axessibility}  

\usepackage{hyperref}

\usepackage{orcidlink}

\usepackage{amsmath,amsfonts,bm}

\def\eqref#1{equation~\ref{#1}}

\def\1{\bm{1}}

\DeclareMathAlphabet{\mathsfit}{\encodingdefault}{\sfdefault}{m}{sl}
\SetMathAlphabet{\mathsfit}{bold}{\encodingdefault}{\sfdefault}{bx}{n}

\newcommand{\OURS}{XYZ-IBD}

\newcommand{\crossmark}[1]{\ding{55}}

\definecolor{mypink}{RGB}{255,105,180}

\begin{document}

\title{XYZ-IBD: Benchmarking Robust 6D Object Pose \\
Estimation under Real-World Industrial Complexity}

\titlerunning{XYZ-IBD: Benchmarking Robust 6D Pose Estimation}

\author{
Junwen Huang$^{1,2}$,
Jiaqi Hu$^{1}$,
Peter KT Yu$^{3,4}$,
Slobodan Ilic$^{1}$,\\
Martin Sundermeyer$^{5}$
Benjamin Busam$^{1,2}$
}

\authorrunning{J.~Huang et al.}

\institute{
$^{1}$Technical University of Munich \quad 
$^{2}$Munich Center for Machine Learning (MCML) \quad \\
$^{3}$XYZ Robotics \quad
$^{4}$ROBOX \quad
$^{5}$Google \quad
}
\maketitle

\begin{figure*}[h]
\centering
\includegraphics[width=\textwidth]{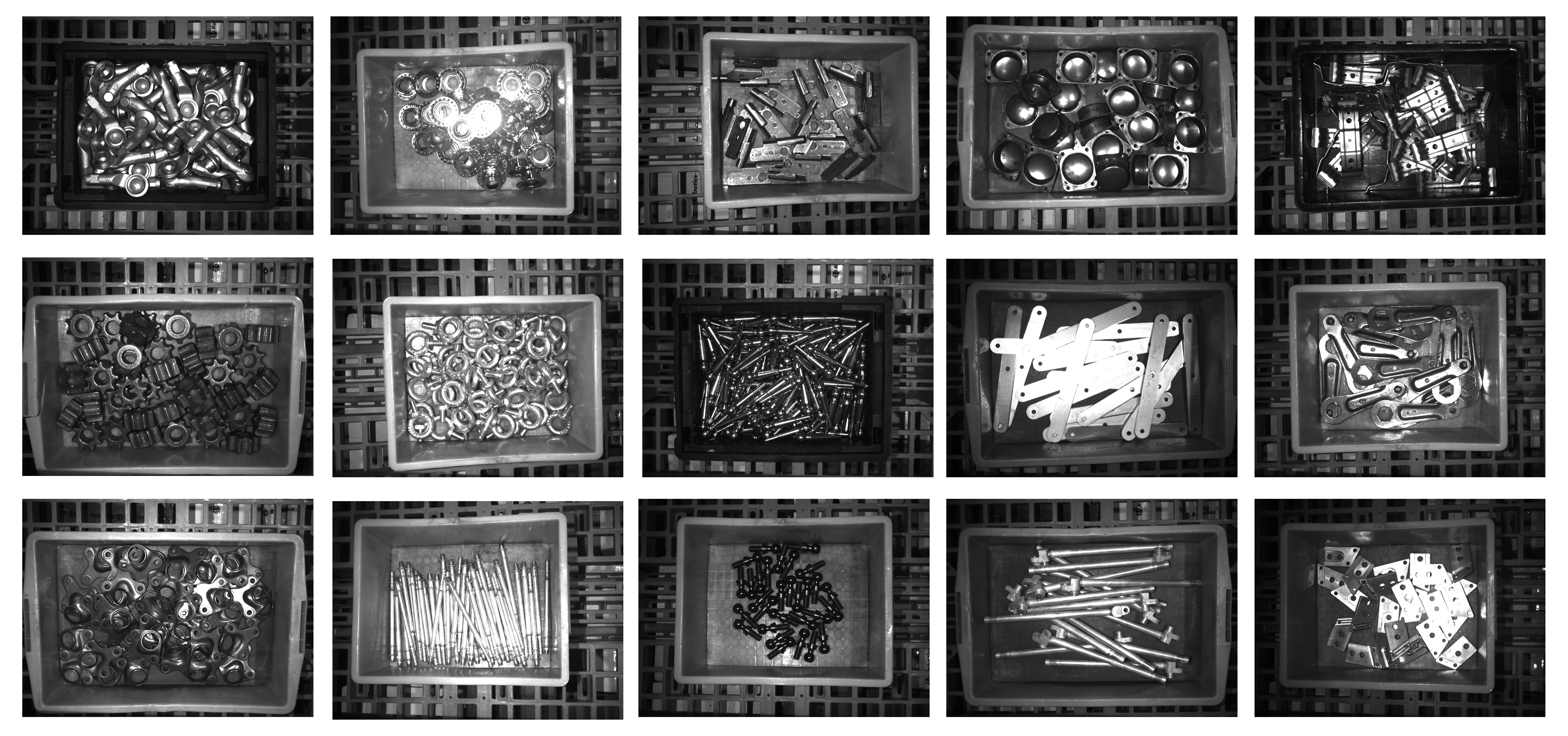}
\caption{ \small XYZ-IBD is an industrial-grade dataset for object detection and 6D pose estimation, providing \textbf{sub-millimeter annotation precision} for reliable evaluation. Unlike household datasets, XYZ-IBD focuses on the perception bottleneck of robotic manipulation: tackling multi-instance ambiguity, randomly stacked scenarios, heavy occlusion, geometric symmetries, and extreme specular reflections. We benchmark the recent SOTA object detection and pose estimation under both object-specific and unseen-object setups, revealing a significant performance gap, challenging existing SOTA methods to bridge the divide between laboratory benchmarks and real-world factory complexity. } 
\label{fig:teaser1} 
\end{figure*}

\begin{abstract}
While current 6D pose estimation benchmarks have reached near-saturation on household objects, they often fail to capture the stochastic and optical complexities of industrial environments. We introduce \OURS{}, a high-precision benchmark for object detection and 6D pose estimation specifically designed for industrial bin-picking. \OURS{} addresses the domain gap by providing 75 multi-view real-world scenes containing approximately 273k annotated instances of metallic, symmetrical, and specular objects. Unlike existing datasets, our benchmark features high-density stochastic stacking and multi-instance ambiguity, reflecting authentic robotic manipulation challenges. We employ a rigorous multi-stage and semi-automatic annotation pipeline, ensuring sub-millimeter annotation accuracy. The annotations are validated through our designed error quantification scheme, securing the reliability of the annotation quality. In addition to real-world evaluation data, we provide a large-scale complementary synthetic training set that is rendered under realistic bin-picking simulation. Benchmarking state-of-the-art (SOTA) methods for 2D detection and 6D pose estimation reveals a significant performance degradation compared to standard household benchmarks, highlighting the unsolved challenges of industrial vision. \OURS{} establishes a new frontier for robust pose estimation in complex, high-occlusion, and reflective scenarios. The dataset and benchmark are publicly available at \url{https://xyz-ibd.github.io}.
\end{abstract}

\section{Introduction}
\label{sec:intro}

Robust 6D object pose estimation is a fundamental prerequisite for autonomous robotic manipulation, particularly in industrial bin-picking~\cite{sundermeyer2023bop}. While state-of-the-art (SOTA) frameworks achieve near-saturation on standard household benchmarks, a critical performance gap emerges in real-world industrial deployment~\cite{van2025bop}. Household datasets~\cite{xiang2018posecnn,kaskman2019homebreweddb,hodan2018bop} typically feature objects with rich textures, semantic cues, and minimal clutter. In contrast, industrial environments present a severe perception bottleneck: parts are often highly reflective, texture-less, and geometrically symmetric~\cite{itodd}. Furthermore, industrial manipulation dictates strict sub-millimeter spatial tolerance; the centimeter-level errors acceptable in household scenarios inevitably lead to catastrophic grasping failures. 

Although recent datasets have introduced texture-less or industrial objects~\cite{hodan2017tless,dimo,liu2021stereobj,ipd,lmo,icbin}, they frequently fail to replicate authentic factory complexity, missing high-density stochastic stacking, intense multi-instance ambiguity, and severe specular reflections. Crucially, obtaining accurate ground truth for highly reflective metallic parts is notoriously difficult. Standard depth sensors suffer from severe noise and geometric distortion on specular surfaces, causing existing industrial datasets to inherit measurement biases that compromise their reliability as rigorous evaluation benchmarks. 

To bridge this divide, we introduce \OURS{}, a novel, high-precision RGB-D benchmark engineered specifically for industrial bin-picking. \OURS{} comprises texture-less, metallic industrial parts arrayed across 75 densely cluttered, multi-view real-world scenes, yielding approximately 273k 6D pose annotations. To ensure robust data modalities, we capture scenes using a multi-sensor suite (structured light, laser scanner, and stereoscopic cameras) mounted on an industrial robotic arm. To guarantee the sub-millimeter precision required for industrial-grade evaluation, we employ a rigorous semi-automatic annotation pipeline. We utilize an anti-reflection coating to suppress specularity~\cite{yang2021robi,jung2023importance,jung2024scrream}, enabling high-fidelity multi-view depth fusion and precise point cloud reconstruction. Furthermore, we explicitly quantify our cumulative annotation error by replicating the physical multi-view setup in a controlled simulation environment. This simulation-to-real validation confirms that our ground truth annotations achieve metrology-grade, sub-millimeter accuracy. We supplement this real-world data with a physics-rendered synthetic dataset of 45k views that fully recap the multi-instance randomness the simulation environment.

Extensive benchmarking of representative SOTA 2D detection and 6D pose estimation frameworks~\cite{wen2024foundationpose,lin2024sam,wang2021gdr} on \OURS{} reveals a stark performance degradation. Models that excel on existing datasets struggle significantly with the visual ambiguity and optical complexity inherent to our scenes. By exposing this domain gap, \OURS{} provides a critical, high-precision testbed to drive the development of robust perception algorithms for industrial automation.

In summary, our key contributions are:
\begin{itemize}
    \item We introduce a challenging real-world industrial bin-picking dataset that captures the extreme complexities of multi-instance stochastic stacking, specular reflection, and geometric ambiguity.
    
    \item We deploy a rigorous semi-automatic annotation pipeline to provide sub-millimeter precision labels suitable for industrial-grade evaluation, which is further verified through a noise-recovered quantification process, securing the reliability of our dataset as a real-world benchmark.  
    
    \item We comprehensively benchmark SOTA instance-specific and generalizable frameworks for object detection and 6D pose estimation, demonstrating a significant performance gap and establishing a new frontier for robust 6D pose estimation.
\end{itemize}
\section{Related Work}
\label{sec:relatedwork}

\subsection{Household Datasets}
A large number of object pose and scene depth datasets have been developed to address everyday scenarios involving household objects. Datasets such as LineMOD~\cite{lm}, LineMOD-Occlusion~\cite{lmo}, YCB-V~\cite{xiang2018posecnn}, HomebrewedDB~\cite{kaskman2019homebreweddb}, and TUD-L~\cite{hodan2018bop} are widely used in benchmarks for model-based object pose estimation~\cite{van2025bop,hodan2018bop}, and have driven progress on key challenges such as handling texture-less objects~\cite{lm}, occluded targets~\cite{lmo}, and typical household environments~\cite{xiang2018posecnn, kaskman2019homebreweddb}. The HOPE dataset~\cite{hope} extends this focus to robotic manipulation scenarios with varied lighting and occlusion conditions. IC-BIN~\cite{icbin} introduces an early bin-picking setup with randomly placed objects, but it includes only two textured objects and suffers from low annotation quality. StereoOBJ-1M~\cite{liu2021stereobj} improves annotation precision through structure-from-motion (SfM) with checkerboards and offers a large number of RGB images, yet it lacks object diversity and does not include depth data. NOCS~\cite{wang2019normalized} presents the first category-level 6D pose dataset, covering six household object categories. More recent datasets such as PhoCal~\cite{wang2022phocal}, HouseCat6D~\cite{jung2022housecat6d}, Booster~\cite{ramirez2023booster}, and SCRREAM~\cite{jung2024scrream} focus on more complex scenes involving transparent or highly reflective objects and utilize a range of sensor modalities, including RGB, depth, and polarization images. While those datasets provide high-quality depth and pose annotations, they lack typical scene properties found in industrial environments. Therefore, existing datasets featuring household objects do not fully capture the challenges inherent in industrial applications, which involve both object-level and scene-level complexity.
\begin{table*}[t]
\centering
\caption{\small Comparison of datasets for object pose estimation from different dimensions.}
\resizebox{\textwidth}{!}{
\begin{tabular}{ccccccccccccc}
\toprule
\textbf{Dataset} & 
\makecell{\textbf{Modalities}} & 
\makecell{\textbf{Number of}\\\textbf{Objects}} & 
\makecell{\textbf{Object}\\\textbf{Diversity}} & 
\makecell{\textbf{Object Diameter}\\\textbf{(mm)}} & 
\textbf{Frames} & 
\textbf{Instances} & 
\makecell{\textbf{Instances}\\\textbf{ per Scene}} & 
\makecell{\textbf{Accurate}\\\textbf{Depth GT}} & 
\textbf{Occlusion} & 
\textbf{Reflection} & 
\makecell{\textbf{Labeling Error}\\\textbf{(mm)}} \\ 
\midrule \midrule
\makecell{DIMO \\ \scriptsize{\cite{dimo}}} & RGB-D & 6  & +   & 75$\sim$302 & 31.2k  & 100k   & <10 & \ding{55} & +   & \ding{55} & 2.7 \\
\makecell{T-LESS \\ \scriptsize{\cite{hodan2017tless}}} & RGB-D & 30 & +++ & 63$\sim$152 & 147k & 100k   & <10 & \ding{55} & ++  & \ding{55} & 11.3 \\
\makecell{ITODD \\ \scriptsize{\cite{itodd}}} & RGB-D & 28 & +++ & 24$\sim$270 & 800   & 5k    & <10 & \ding{55} & ++  & \ding{55} & 1.8 \\
\makecell{ROBI \\ \scriptsize{\cite{yang2021robi}}} & RGB-D & 7  & +   & 24$\sim$76  & 8k    & 600k  & >10 & \checkmark & +++ & \checkmark & 1.8 \\
\makecell{StereOBJ-1M \\ \scriptsize{\cite{liu2021stereobj}}} & RGB & 18 & ++  & -   & 396k  & 1.5M  & <10 & \ding{55} & +   & \ding{55} & 2.3 \\
\makecell{IPD \\ \scriptsize{\cite{ipd}}} & RGB-D+Polar & 20 & +++ & 80$\sim$240 & 30k   & 100k  & <10 & \ding{55} & +   & \checkmark & N/A \\
\makecell{XYZ-IBD (Ours)} & RGB-D & 15 & +++ & \textbf{54$\sim$300} & 22.5k & 273k & \textbf{22} & \checkmark & \textbf{+++} & \checkmark & \textbf{0.99} \\
\bottomrule
\end{tabular}
}
\vspace{-0.5cm}
\label{tab:dataset_comparison_combined}
\end{table*}

\subsection{Industrial Datasets}
In industrial applications, the working environment is quite different from the household scenario. Firstly, unlike household objects, industrial parts are usually texture-less and often symmetric and highly reflective~\cite{itodd,ipd,yang2021robi}. Consequently, networks trained on household objects hardly generalize to industrial datasets. Secondly, the required pose accuracy in industrial robotics is usually higher than in household robotics or AR/VR applications. The robotic arm is expected to not only pick up singulated objects, but typically needs to pick objects from a filled container and place them at a target pose or assemble them. Even though bin-picking is a typical setup for industrial applications, only a few publicly available datasets target this scenario which severely hampers the usability of pose estimation pipelines in industrial practice.
T-LESS dataset~\cite{hodan2017tless} features texture-less industrial objects with symmetries but does not present challenging lighting conditions, and the annotation quality is not mm-accurate, a requirement in many industrial applications. Only a few scenes present the complexity of real bin-picking configurations where similar objects occlude each other. 
ITODD~\cite{itodd} collects industrial parts with challenging geometry and lighting conditions but does not feature bins filled with objects. The consistently low pose estimation scores on ITODD~\cite{itodd} in the BOP challenge~\cite{van2025bop} also demonstrate the need for industrial bin-picking datasets. 
Other datasets such as DIMO~\cite{dimo} and ROBI~\cite{yang2021robi} focus on metallic objects for bin-picking setups, but they focus on a limited number of objects whose size and shape are not representative of the diversity in real applications. 
The recent dataset IPD~\cite{ipd} leverages multiple sensors to collect data from industrial objects but presents little clutter, stacking and occlusions which simplifies the setup compared to real industrial scenarios. Table~\ref{tab:dataset_comparison_combined} compares the characteristics of current industrial datasets.

\section{The \OURS~Dataset}

\begin{figure*}[t]

\centering
\includegraphics[width=\textwidth]{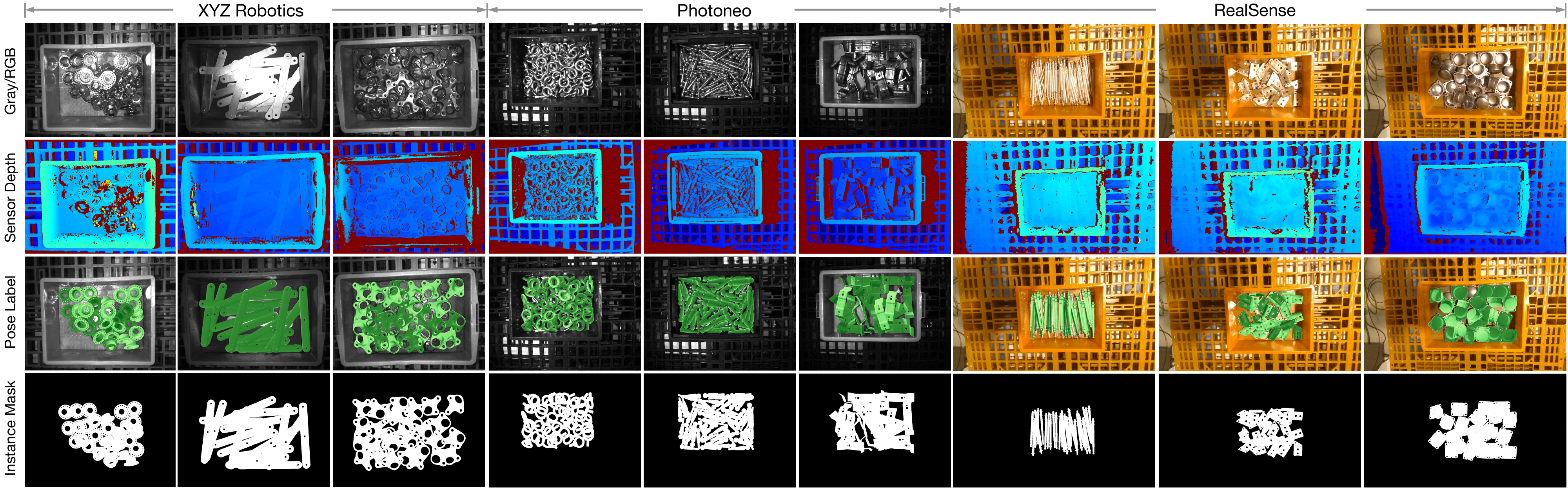}
\caption{\small \textbf{Multi-sensor Data and High-Precision Annotations in \OURS{}.} Synchronized RGB, grayscale, and depth streams acquired via a tri-sensor suite capture severe industrial perception bottlenecks. The scenes feature geometrically ambiguous, texture-less instances under complex illumination, stochastically stacked to induce high-density clutter and severe occlusion. Overlaid are our high-fidelity 6D pose and instance mask annotations, demonstrating the sub-millimeter precision required for rigorous industrial benchmarking.}
\label{fig:teaser} 
\end{figure*}
As shown in Figure \ref{fig:teaser},
\OURS~establishes a benchmark for industrial bin-picking by capturing data under authentic factory conditions. It advances prior work through four perspectives:
\textbf{(1) Industrial-Grade Setup}: Data is acquired using industry-standard robotic arms (FANUC M10iD/8L) and multi-modal sensors (RGB/depth/grayscale) mounted at industrial working distances, replicating real application conditions.
\textbf{(2) Challenging Objects}: fifteen reflective, texturless, and mostly symmetric industrial parts that present rich geometrical shapes and sizes(54–300 mm scale), introducing academic challenges for pose estimation. 
\textbf{(3) Multi-instance Dense Clutter}: Objects are randomly and densely arranged in a container with multiple repeat instances, creating more ambiguity for instance detection and alignment.
\textbf{(4) Precise Annotation}: Our annotation pipeline achieves $<$1 mm positional and $<$1° angular annotation accuracy, validated with simulated environment.

This benchmark consists of \textbf{75 real-world scenes} (five configurations per object), encompassing approximately \textbf{273k annotated instances} with \textbf{22 instances per scene on average}, and \textbf{up to 60 instances in some scenes}. In addition, it includes \textbf{45k synthetic training set} generated with BlenderProc~\cite{denninger2020blenderproc} through physics-based object interactions, simulating a realistic bin-picking setup.

\subsection{Objects and Hardware}
\subsubsection{Objects and Instance Distribution.}
\begin{figure*}[t]
\centering
\includegraphics[width=\textwidth]{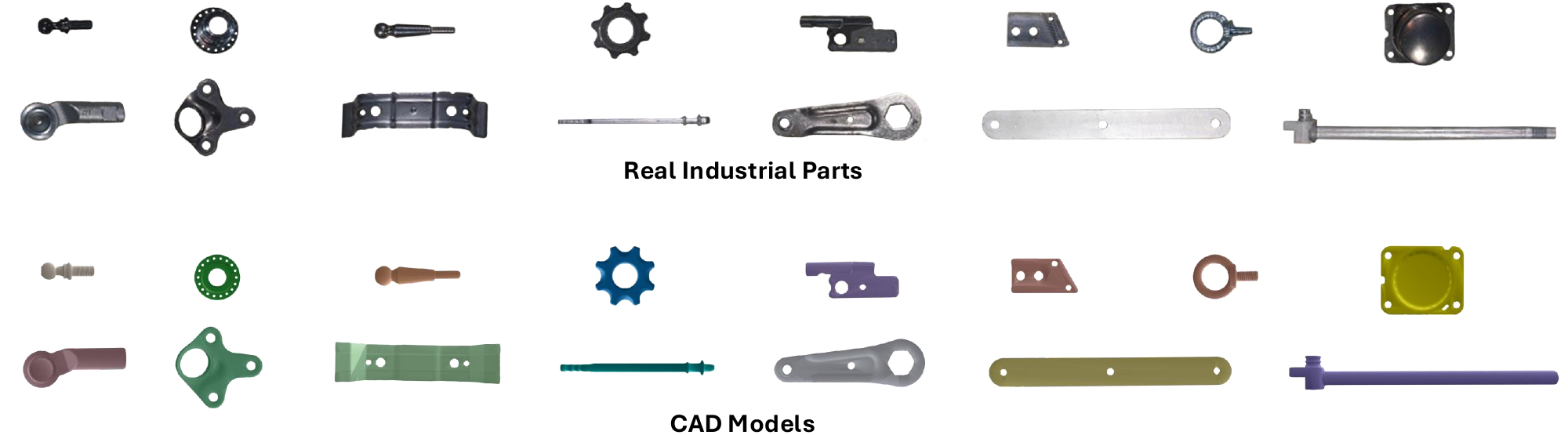}   
\caption{The real industrial parts.} 
\label{fig:objects}
\end{figure*}

Our dataset comprises fifteen representative industrial parts with diameters ranging from 54~mm to 300~mm, including components like sheet metal parts, bolts, pins, covers, and many other kinds of machined metal objects. As shown in Figure~\ref{fig:objects}, these objects exhibit challenging visual properties such as high reflectivity and symmetry that are common in manufacturing environments yet problematic for vision algorithms. The original CAD models provided by industrial partners ensure micron-level geometric accuracy for both real-world captures and synthetic renderings. All real-world data is collected in bins with sensor-to-object distances carefully calibrated between 600 and 1000mm. We randomly stacked multiple instances of each object into the bin, with severe occlusion and clutter. Figure~\ref{fig:instance_distribution} illustrates the distribution of instance counts across different datasets. Compared to existing industrial object datasets, ours exhibits the highest density of visible objects per image. This high density specifically addresses key challenges in object detection and pose estimation, particularly in scenarios characterized by significant visual ambiguity.

\subsubsection{Sensor Setup.}
\begin{figure*}[h]
\centering
\begin{minipage}[c]{0.5\textwidth}
    \centering
    \includegraphics[width=\textwidth]{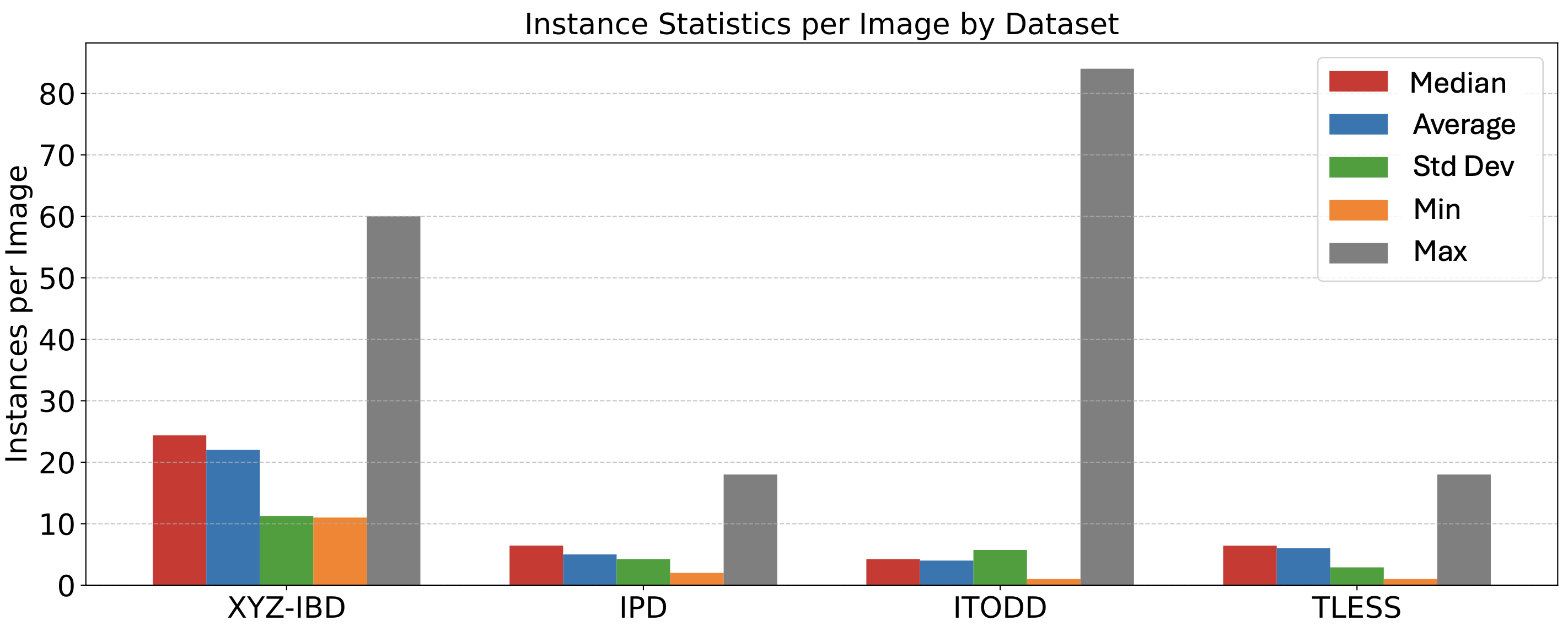}
    \caption{\small Comparison of per-image instance distribution among different industrial datasets. Our dataset presents the highest number of instances per image, introducing challenges of visual ambiguity.}
    \label{fig:instance_distribution}
\end{minipage}
\hfill
\begin{minipage}[c]{0.48\textwidth}
    \centering
    \includegraphics[width=\textwidth]{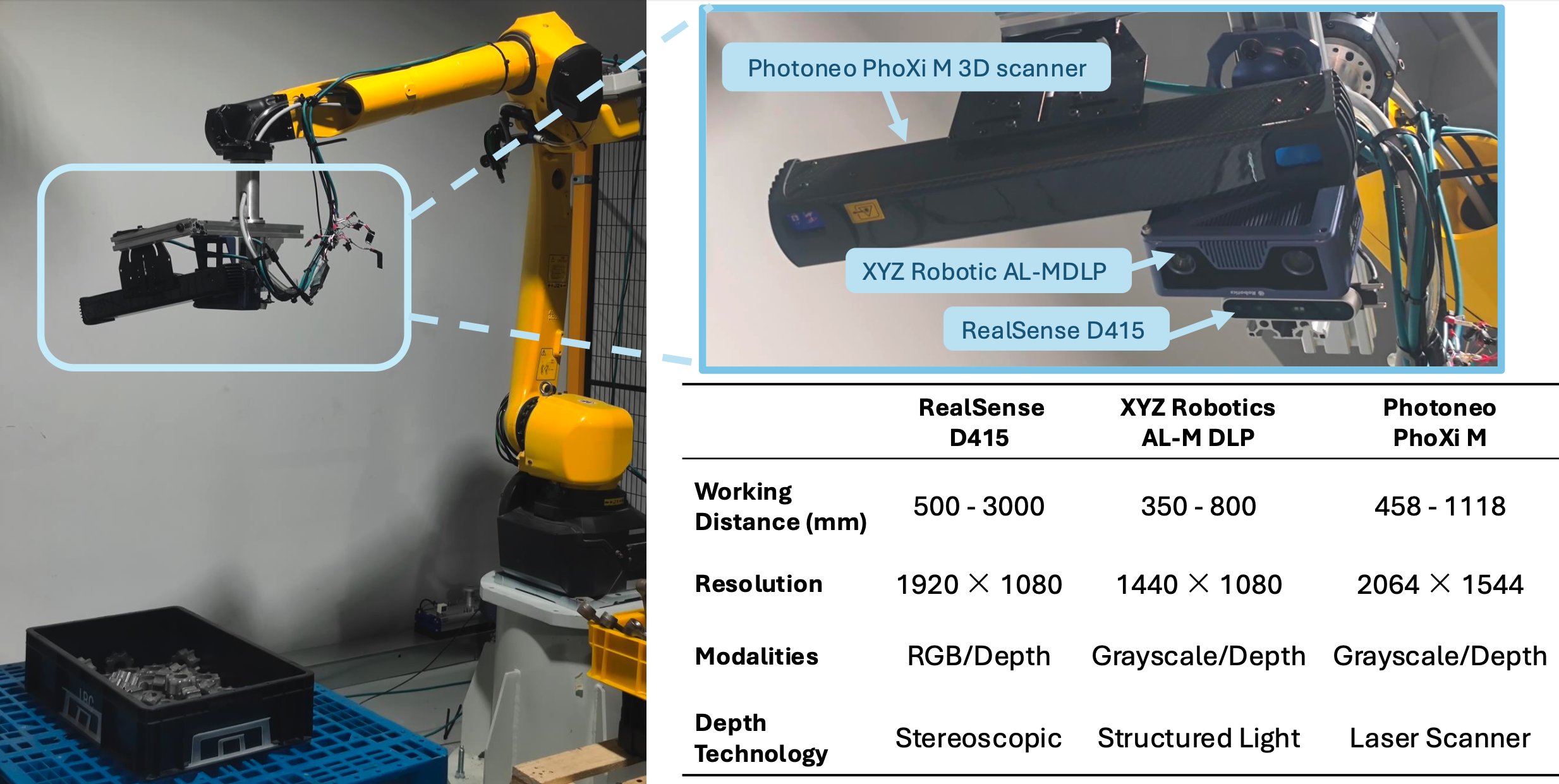}
    \caption{\small The robotic setup and sensor parameters for data collection, following the real-world factory-level working layout.
}
    \label{fig:cam_setup}
\end{minipage}
\end{figure*}

For precise and repeatable data acquisition, we employ an industrial-grade FANUC M10iD/8L robotic arm with ±0.06mm repeatability to position our multi-sensor array. Three complementary vision systems are rigidly co-mounted on the end-effector (see Figure~\ref{fig:cam_setup}):
the \textit{Intel RealSense D415 stereoscopic camera} provides aligned RGB (1920×1080) and depth streams at 30~FPS, offering baseline color-depth registration for general scene understanding;
the \textit{XYZ Robotic AL-M DLP structured-light} camera delivers high-precision grayscale (1440×1080) and depth maps (0.08mm resolution) through projected pattern deformation analysis, particularly effective for matte surfaces;
the \textit{Photoneo PhoXi M 3D scanner} utilizes laser triangulation to generate high-accuracy depth data (up to 2064×1544 resolution) with 0.1~mm voxel precision, complemented by synchronized grayscale imagery.
All sensors are positioned at optimized working distances of 600-1000~mm based on object size and bin geometry, maintaining consistent fields-of-view across the industrial container. The fixed relative positions between cameras enable direct cross-modality calibration, while the robotic arm's precise positioning ensures reproducible viewpoint acquisition throughout the data collection process.

\subsection{Data Acquisition Pipeline.}
\begin{figure*}[t]
    \centering
    \includegraphics[width=1\textwidth]
    {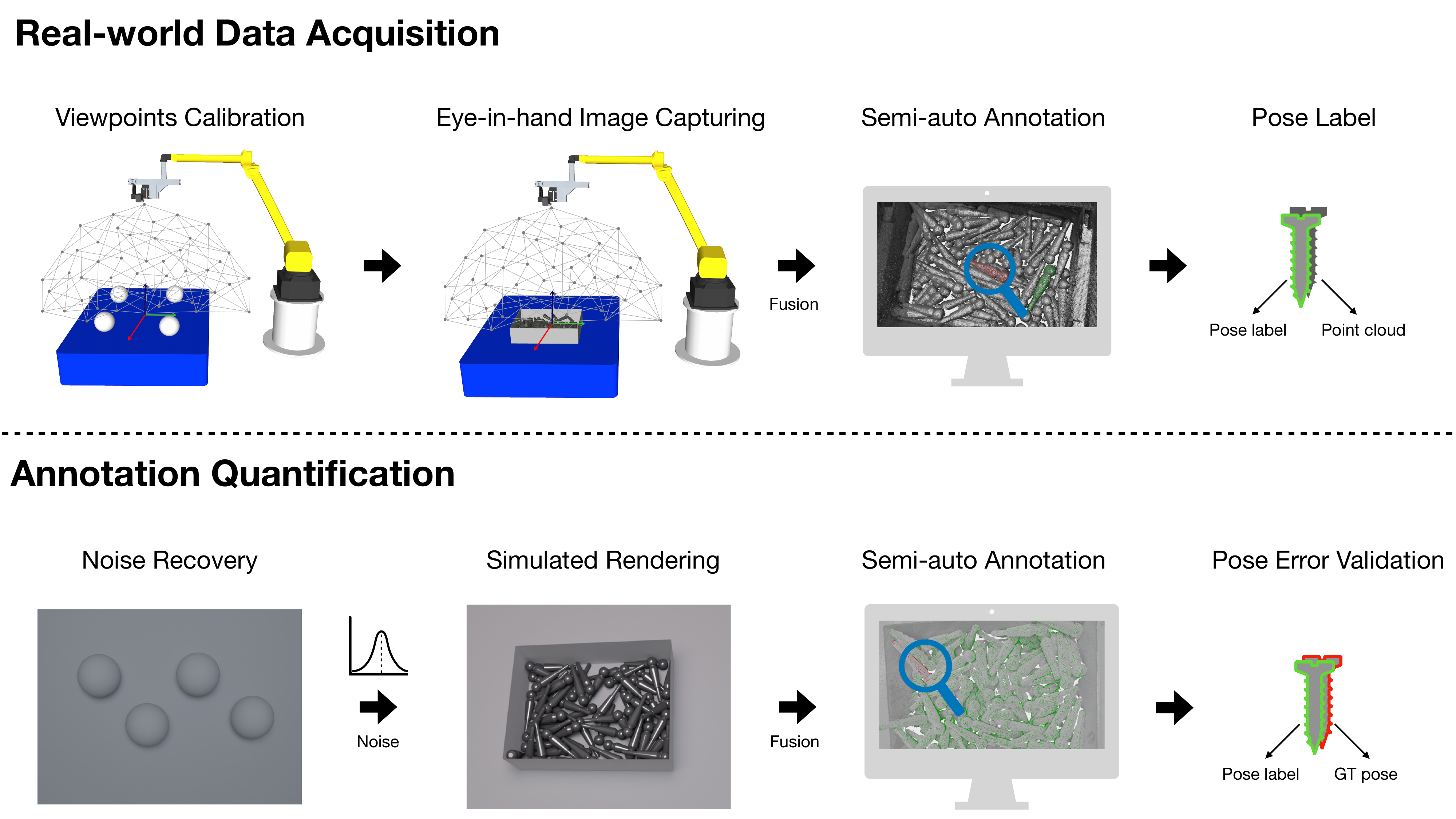}
    \caption{\small The real-world industrial data collection pipeline (upper) and the annotation error quantification pipeline in the simulated environment(lower). We deploy a semi-auto pipeline to ensure a high-precision annotation protocol. The annotations are validated through our designed error quantification scheme, securing the reliability of the annotation quality.}
    \label{fig:overview}
\end{figure*}

As shown in Figure \ref{fig:overview}, our data acquisition pipeline integrates three sequential stages: viewpoint sampling and calibration, multi-pass scene capture for depth ground truth, and a hybrid annotation protocol combining manual and algorithmic refinement.

\subsubsection{Multi-view Sampling and Calibration.}
Beginning with the bin’s centroid as the origin, we define a spherical sampling surface spanning elevation angles of 45° to 90° to balance perspective diversity and robotic arm operability. Fifty viewpoints are randomly distributed across this surface to ensure comprehensive spatial coverage. Following the calibration framework of \cite{yang2021robi}, we place four precisely machined calibration spheres on the working plane. During an initial calibration pass, the robotic arm captures multi-modal images of these spheres across all viewpoints. The cameras are firstly undistorted and obtain the initial camera poses with hand-eye calibration~\cite{handeye}, then pose refinement via iterative closest point (ICP) alignment on the spheres’ point clouds establishes relative transformations between viewpoints with around 0.248 mm average root mean square error (RMSE), resolving the 6~DoF relationships between 49 secondary viewpoints and a primary reference view. These transformations enable subsequent multi-view data fusion and label projection with sub-millimeter consistency. After calibration, the calibration spheres are removed, and the robotic arm systematically revisits each pre-calibrated viewpoint to capture cluttered industrial bin-picking scenes. At each viewpoint, three rigidly mounted cameras (Intel RealSense D415, XYZ Robotic DLP, Photoneo PhoXi) acquire synchronized RGB, grayscale, and depth data. To maximize scene diversity, we perform five complete capture cycles per object, randomly shuffling parts between cycles.

\subsubsection{Multi-Pass Scene Capture.}
To address depth sensing challenges from reflective surfaces, we employ a dual-phase capture strategy. The first phase applies a temporary anti-reflective coating (Acksys SP-102) to suppress specularity, enabling high-fidelity ground truth depth acquisition. After allowing 15 minutes for complete evaporation under controlled ambient conditions (25°C ±1°C), we execute an identical second capture pass to record the scene’s native optical properties. Both phases maintain pixel-wise spatial correspondence through robotic arm pose repetition (±0.06~mm precision), providing high-quality depth to fuse the scene point cloud, as shown in  Figure \ref{fig:depth_vis}, thus resulting in more accurate pose annotation and also aligned datasets of enhanced and raw depth for algorithm benchmarking. 

\subsubsection{6D Pose Annotation.}
The annotation derives from a hierarchical process beginning with 3D fusion of spray-enhanced depth data into a unified scene point cloud. Annotators coarsely align CAD models to this reconstruction using our developed constrained GUI (±1 mm translational, ±1° rotational increments), followed by multi-scale ICP refinement. The ICP pipeline first aligns downsampled point clouds for global adjustment and then iteratively optimizes with full-resolution data to achieve sub-millimeter accuracy. Finalized poses are propagated to all 50 viewpoints using pre-calibrated transformations that we obtained from the first stage, ensuring label consistency across perspectives without manual per-view annotation. This protocol yields around 273k annotated object instances in total, with a minimum of 11 and a maximum of 60 instances per scene, resulting in an average of 22 instances per image. We choose the XYZ Robotic DLP camera as the primary camera and perform annotations on the data it collected. The annotations of the other two cameras are projected through the calibrated relative transformation between the cameras.  

\subsection{Annotation Error Quantification.}
As illustreated in Figure \ref{fig:overview}, to quantify the cumulative annotation error, we replicated the data collection and annotation process in a simulated environment, recovering the sensor error, calibration error and the human annotation error, and compared the resulting annotations against ground truth poses. Our evaluation framework comprises three stages: data noise recovery, simulated data collection, and 6D pose quantification.

\begin{figure*}[t]
\centering
\begin{minipage}[c]{0.5\textwidth}
    \centering
    \includegraphics[width=\textwidth]
    {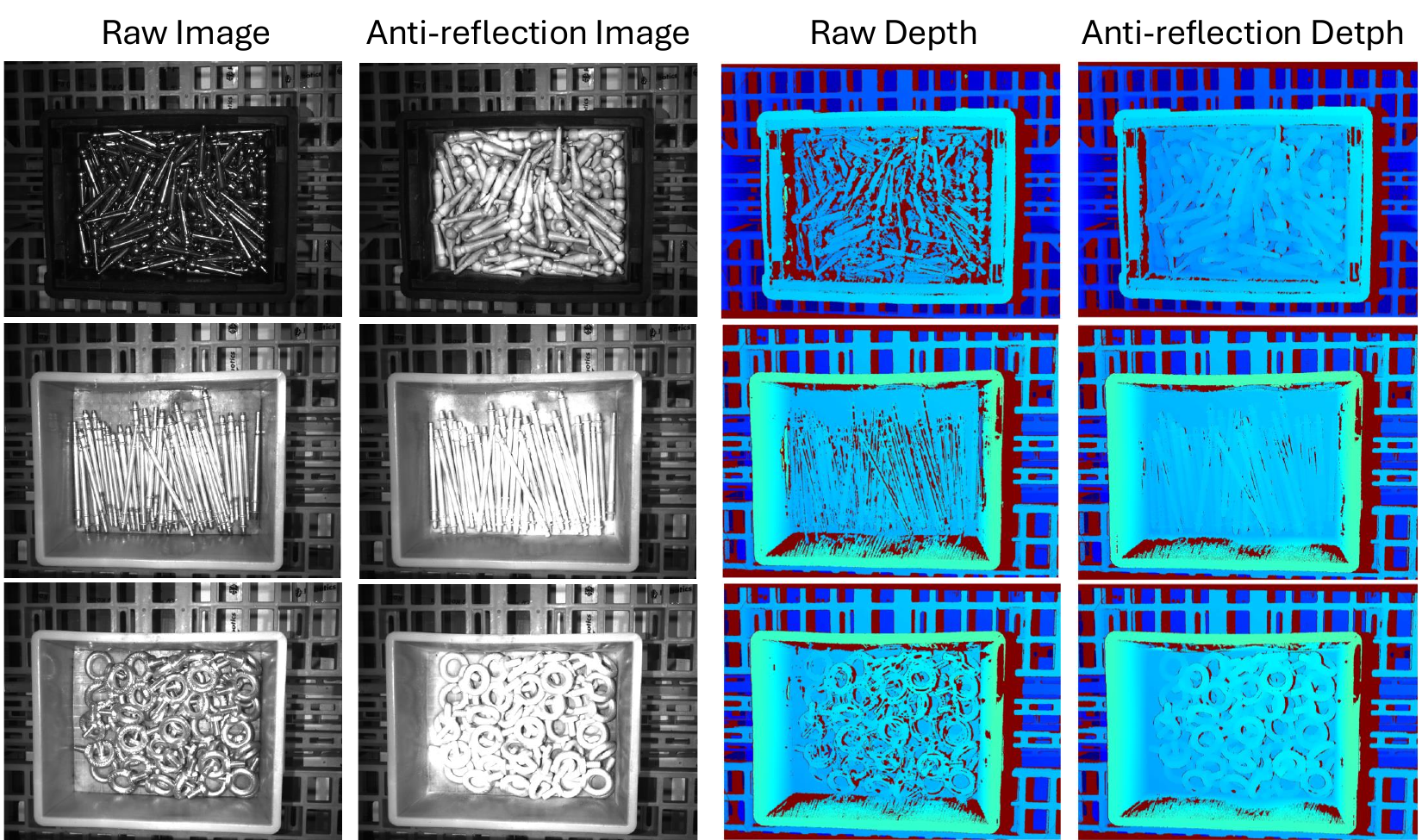}
    \caption{\small Comparison between raw depth and anti-reflection depth. 
}
    \label{fig:depth_vis}
\end{minipage}
\hfill
\begin{minipage}[c]{0.48\textwidth}
    \centering
    \includegraphics[width=\textwidth]{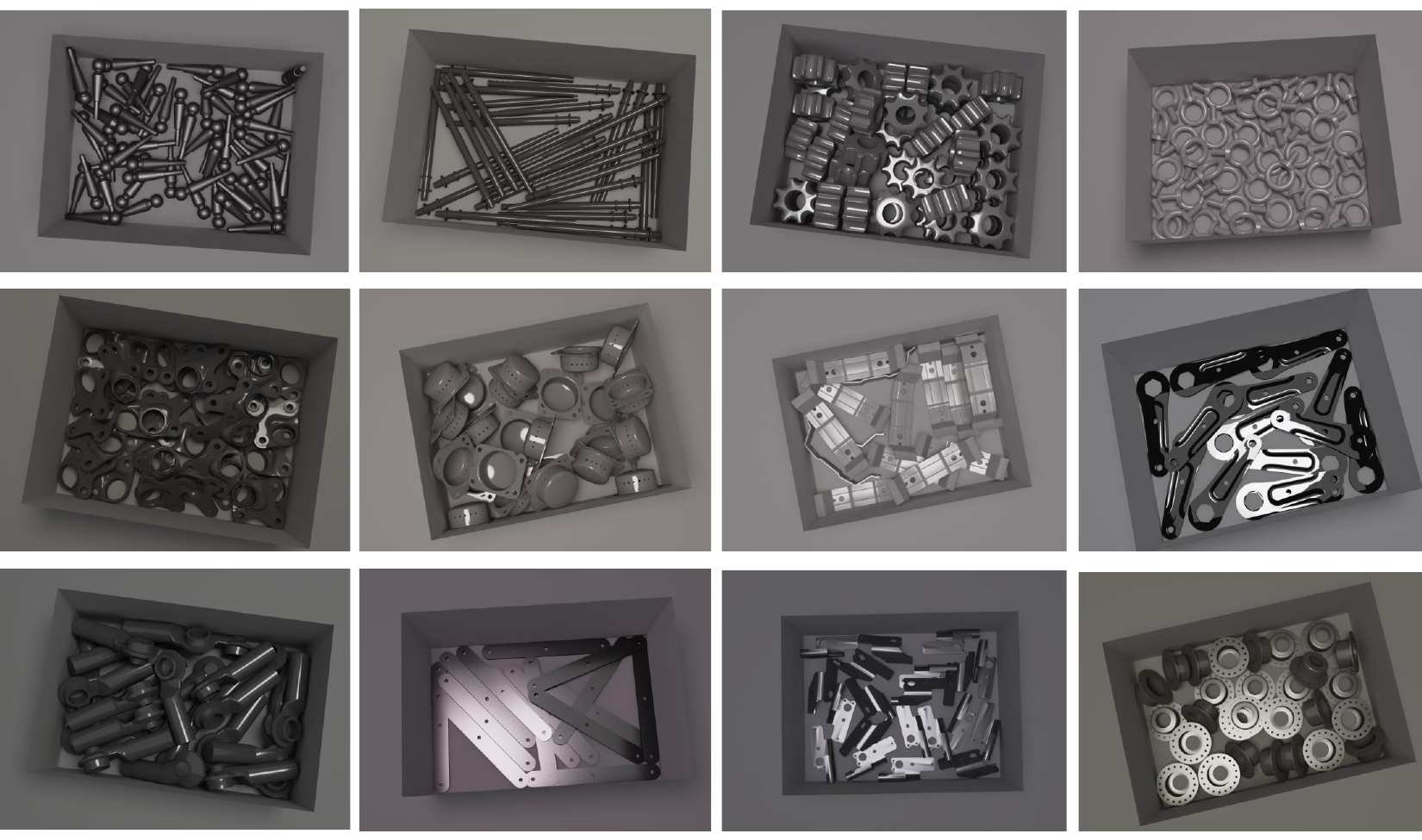}
    \caption{\small The complementary synthetic training images. 
    }
    \label{fig:syn_vis}
\end{minipage}
\end{figure*}

\subsubsection{Data Noise Recovery.}
We simulate the multi-view calibration procedure in a synthetic environment using identical calibration spheres and camera parameters as in the real-world setup. 50 calibration views are sampled, and varying levels of Gaussian noise are added to the rendered depth images. The corresponding RMSE of the point cloud is then computed using ICP, revealing the relationship between noise magnitude and calibration error (see Table~\ref{tab:rmse}). When Gaussian noise $\sigma = 0.26$~mm is applied, the computed RMSE reached 0.248 mm, matching the error observed during real-world calibration. This provided the chosen noise level to best represent the cumulative error introduced by the sensor, robotic system, and multi-view calibration process.

\subsubsection{Simulated Data Collection.}
To ensure realistic, cluttered arrangements, we generate synthetic counterparts using the same CAD models within a simulated bin-picking environment rendered with physically-based rendering in BlenderProc~\cite{denninger2020blenderproc}. Objects are randomly dropped into the bin via a free-fall simulation, and any that fall outside the bin are removed. Multi-view synthetic images are rendered using a complementary noise model derived in the last step. The same annotation pipeline used for real data, incorporating multi-view fusion, manual adjustments, and multi-scale ICP refinement, is also applied to the synthetic scenes. As ground truth poses are available in the simulation, this setup allows for direct comparison between annotated and true object poses.

To complement the real dataset, we additionally render a large-scale synthetic dataset as the training data. We programmatically vary rendering conditions, including lighting, material properties, object quantity, and pose configurations, closely replicating real-world setups to ensure cross-domain consistency. For each of the 15 objects, we perform 120 free-fall simulations, with random variations in material and lighting, resulting in a total of approximately 45,000 frames in the synthetic training dataset. As shown in Figure~\ref{fig:syn_vis}, this parallel real-synthetic collection supports robust benchmarking while preserving strong visual alignment between domains.

\begin{table*}[t]
\centering
\begin{minipage}[t]{0.55\textwidth}
\centering
\caption{\small Specifications for data collection error sources.}\vspace{-0.2cm}
\resizebox{\textwidth}{!}{
\begin{tabular}{llc}
\toprule
\textbf{Source} & \textbf{Specify} & \textbf{Error (mm)} \\
\midrule
\multirow{5}{*}{\shortstack{Viewpoints\\Calibration}} 
                        & Sensor Calibration & 0.10  \\
                        & Sensor Temporal noise & 0.10      \\
                        & Sensor Distortion & N/A           \\
                        & Robot arm Repeatability & 0.06     \\
                        & Viewpoints Calibration RMSE (Total) & \textbf{0.245}  \\
\midrule
Depth fusion & TSDF & N/A\\
Manually annotate & Human, ICP & N/A\\
\midrule
Overall &  & \textbf{0.99}\\
\bottomrule
\end{tabular}
}
\label{tab:errors}
\end{minipage}
\hfill
\begin{minipage}[t]{0.4\textwidth}
\centering
\caption{\small Effect of Gaussian noise on average viewpoints calibration RMSE.}\vspace{-0.2cm}
\resizebox{\textwidth}{!}{
\begin{tabular}{cc}
\toprule
\textbf{\shortstack{Gaussian Noise $\sigma$ (mm)}}    & \textbf{\shortstack{RMSE (mm)}}\\ 
\midrule
0.0   & 0.199       \\ 
0.1   & 0.209       \\ 
0.2   & 0.233       \\ 
\textbf{0.26}   & \textbf{0.248}       \\ 
0.3   & 0.259       \\ 
\bottomrule
\end{tabular}
}
\label{tab:rmse}
\end{minipage}
\vspace{-5mm}
\end{table*}

\subsubsection{6D Pose Quantification.}
To assess annotation quality, we systematically investigated several error sources: inherent sensor inaccuracies, robotic arm repeatability, viewpoint calibration discrepancies, and annotator subjectivity. Specifically, the sensor error encompasses both the camera calibration inaccuracies and measurement noise due to sensor characteristics and environmental conditions. This, together with the robot arm repeatability, is manifested in the overall multi-view pose calibration error. We compute pose errors through nearest-neighbor matching between annotated and GT poses using Hungarian assignment on 3D centroid distances. We analyze up to 60 samples per scene $\times$ 3 scenes per object, revealing a mean positional error of 0.999~mm ($\sigma = 0.12$~mm) and an angular error of 0.432$^\circ$ ($\sigma = 0.08^\circ$).  
Per-object error averages across 15 industrial parts demonstrate sub-millimeter precision even for challenging geometries. This synthetic validation confirms that our real-world annotations achieve $<$1~mm positional and $<$1$^\circ$ angular accuracy relative to physical GT.

\section{Benchmarks}
\label{sec:benchmarks}

\subsection{Dataset Split}
\subsubsection{Synthetic Training set.} As showcased in Fig.~\ref{fig:syn_vis}, the synthetic training dataset is generated using high-fidelity CAD models of the industrial components. To simulate realistic bin-picking scenarios, we employ a physics engine to model the free-fall and stochastic settling of multiple object instances range from 10 to 60. For each of the 120 generated scenes per object, we render 25 frames under diverse illumination conditions and varying material properties, utilizing camera intrinsics that match the XYZ structured-light sensor. This process yields approximately 3,000 frames per object, totaling 45k RGB-D samples. The resulting dataset provides comprehensive ground truth, including instance masks, depth maps, and 6D poses, facilitating supervised training for depth estimation, 2D object detection, and 6D pose estimation tasks.

\begin{figure}[t]
\centering
\includegraphics[width=\textwidth]{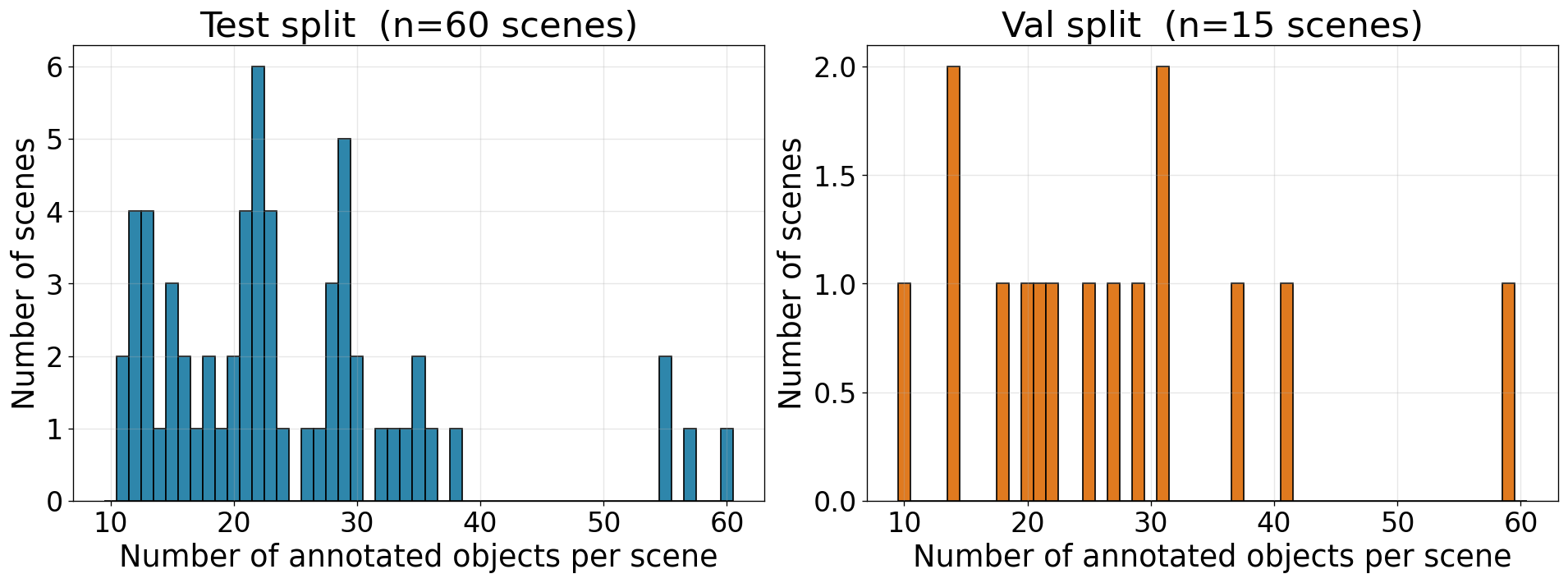}
\caption{ \small The distribution of per-scene instance density in the XYZ-IBD dataset's test and validation splits. } 
\label{fig:distribution} 
\end{figure}
 
 \subsubsection{Real-world Validation and Testing and Set.} For each industrial part, we captured five distinct scenes by systematically varying instance counts, stacking configurations, and illumination conditions with 3 above mentioned type of cameras. Among the 75 real-world scenes, 15 scenes with 2250 different frames are designated for the validation set, while the remaining 60 scenes with 9000 different frames are reserved for testing. As the per-scene instance histogram shown in Figure~\ref{fig:distribution}, most scenes contain \textbf{20--30} instances, tailing to \textbf{60}. Figure~\ref{fig:instance_distribution} also indicates that we have the highest mean density among existing industrial datasets.  The achieved sub-millimeter annotation precision ensures reliable evaluation metrics, even within this highly challenging setup.
 The synthetic training dataset comprises approximately 80 GB in size. The validation and test sets occupy roughly 8 GB and 3 GB, respectively. We will publicly release the object CAD models along with the fully annotated training and validation sets. While the test set images will be made public, the corresponding ground truth will be hosted on our evaluation server to facilitate an open leaderboard for benchmarking.

\subsection{Evaluation Setups}
Our dataset provides high-precision 6D object pose and depth annotations, enabling the establishment of a comprehensive benchmark for \textbf{object detection and pose estimation}. In order to align the evaluation protocol with the widely used Benchmark for 6D Object Pose Estimation (BOP) and challenges, we adopt their evaluation protocols to assess performance on our dataset.
For the 2D detection and 6D pose estimation tasks, we evaluate representative methods under both \textbf{seen and unseen object settings}. In the seen object setup, models are trained on our synthetic dataset and evaluated on the real-world test split. In the unseen object setup, we directly use off-the-shelf generalizable methods, which have been pretrained on large-scale external datasets, to infer our real test scenes without extra finetuning. We benchmark several recent state-of-the-art methods across \textbf{five BOP-Core datasets} (LM-O~\cite{lmo}, T-LESS~\cite{hodan2017tless}, YCB-V~\cite{icbin}, IC-BIN~\cite{icbin}, TUD-L~\cite{hodan2018bop}) in compare with XYZ-IBD, under \textbf{four tasks }(model-based seen/unseen object 2D detection and model-based seen/unseen object 6D detection). All seen-object baseline methods are trained on our synthetic training dataset and evaluated on the real testing split. For all the unseen baselines, we directly use the pretrained model to infer on the testing split. 

\subsection{Evaluation Criteria}
\subsubsection{Object 2D Detection Metics.}
For the object 2D detection task, we follow the model-based 2D detection task defined in BOP 2024-2025 Challenge~\cite{van2025bop}. The objective is to generate a set of non-overlapping 2D binary instance masks with associated confidence scores from an RGB-D input image that contains multiple object instances from a given dataset. To evaluate performance, we adopt the Average Precision (AP) metric, following the protocol used in the COCO 2020 challenges~\cite{coco}. AP is calculated by averaging the precision scores at several Intersection-over-Union (IoU) thresholds, ranging from 0.5 to 0.95 in increments of 0.05. Each object’s AP score reflects its detection quality across these thresholds. To obtain an overall dataset-level performance measure, the mean Average Precision (mAP) is computed by averaging the AP scores across all object categories. This evaluation strategy comprehensively captures both the accuracy of object localization and the effectiveness of category-level recognition, ensuring alignment with established benchmarking standards.

\subsubsection{Object 6D Detection Metics.}
For the 6D pose estimation task, we adopt the model-based 6D object detection metric defined in BOP 2024-2025 Challenge~\cite{van2025bop}, evaluating detection accuracy using symmetry-aware Average Precision (AP) scores. For each predicted pose $\hat{P}$ and its corresponding ground truth pose $P_{GT}$, we compute two error metrics: \textit{Maximum Symmetry-Aware Surface Distance (MSSD)} and \textit{Maximum Symmetry-Aware Projection Distance(MSPD)}. MSSD measures the maximum 3D surface deviation under object symmetries, defined as $e_{\text{MSSD}} = \max_{x \in M} \min_{S \in S} | \hat{P}x - S(P_{GT}x) |$, where $M$ is the object mesh and $S$ is the set of predefined symmetry transformations. MSPD evaluates the maximum 2D projection deviation considering object symmetries, computed as $e_{\text{MSPD}} = \max_{u \in U} \min_{S \in S} | \Pi(\hat{P}x_u) - \Pi(S(P_{GT}x_u)) |$, where $\Pi$ denotes the camera projection function and $U$ the set of visible mesh vertices. A pose estimate is deemed correct when the error $e$ falls below a threshold $\theta_e$. For each error type $e \in {\text{MSSD}, \text{MSPD}}$ and object $o \in O$, we compute the object-level AP score as $AP_{e,o} = \frac{1}{|\Theta_e|} \sum_{\theta \in \Theta_e} P_o(\theta)$, where $\Theta_e$ is the set of threshold values and $P_o(\theta)$ is the precision at threshold $\theta$. The final AP score aggregates over all objects and both error types as $\text{AP} = \frac{1}{2|O|} \sum_{o \in O} \sum_{e \in {\text{MSSD}, \text{MSPD}}} AP_{e,o}$.

\begin{table}[t]
\centering
\caption{\small Performance comparison of 2D and 6D detection SOTA methods on previous core datasets in BOP and XYZ-IBD dataset. We report the AP scores on the main BOP datasets YCB-V, T-LESS, and LM-O, and the average AP on the 5 BOP core datasets (YCB-V, T-LESS, LM-O, IC-BIN and TUD-L) under the four tasks of seen object 2D detection, unseen object 2D detection, seen object 2D detection, and unseen object 2D detection.} 
\label{tab:pose_baselines}
\resizebox{\textwidth}{!}{
\begin{tabular}{l|c|c|c|c|c|c|c}
\toprule
\textbf{Method} & \textbf{Task} & \textbf{Unseen Object} & \textbf{YCB-V} & \textbf{T-LESS} & \textbf{LM-O} & \textbf{BOP-Core 5} & \textbf{XYZ-IBD} \\
\midrule
YOLOX~\cite{ge2021yolox}        & \multirow{4}{*}{2D Detection}  & \ding{55} &0.877 &0.707  &0.894  &0.798 & 0.774 \\
CNOS(SAM)~\cite{nguyen2023cnos}      &    &\checkmark &0.490  &0.395	&0.330   &0.361  & 0.275 \\
SAM-6D(SAM)~\cite{lin2024sam}        &    &\checkmark &0.518  &0.465	&0.437   &0.449  & 0.296 \\
NIDS-Net~\cite{lu2025adapting} &    &\checkmark &0.650  &0.496	&0.439   &0.494  & 0.258 \\
\midrule
GDRN~\cite{wang2021gdr}           & \multirow{5}{*}{6D Detection} & \ding{55} & 0.906 & 0.852 & 0.775 & 0.827 & 0.266 \\
SurfEmb~\cite{haugaard2022surfemb} &  & \ding{55} & 0.799 & 0.828 & 0.760 & 0.758 & 0.247 \\

MatchU~\cite{huang2024matchu} &  & \checkmark & 0.756 & 0.668 & 0.680 & 0.705 & 0.529 \\
FoundationPose~\cite{wen2024foundationpose} &  & \checkmark & 0.889 & 0.646 & 0.756 & 0.734 & 0.564 \\
SAM6D~\cite{lin2024sam}           &  & \checkmark & 0.845 & 0.515 & 0.699 & 0.704 & 0.578 \\
\bottomrule
\end{tabular}
}
\end{table}

\subsection{Evaluation of Object 2D Detection}

\subsubsection{Seen Object 2D Detection.} YOLOX~\cite{ge2021yolox} is a widely used advanced real-time object detection model that builds upon the YOLO~\cite{yolo} series. We follow the implementation of the SOTA method for object pose estimation GDRNet~\cite{wang2021gdr} to train the YOLOX model on our synthetic training data and test on the real test split.

\subsubsection{Unseen Object 2D Detection.} CNOS~\cite{nguyen2023cnos} is a model-based method that uses vision foundation models SAM~\cite{kirillov2023segment} and DINOv2~\cite{oquab2023dinov2} for novel object segmentation and detection without re-training. It renders object templates from a CAD model and ranks SAM-generated segments by comparing their DINOv2 class token features with those of the templates. NIDS-Net~\cite{lu2025adapting} uses a similar workflow but takes Grounding DINO~\cite{liu2024grounding} as the backbone. 
SAM-6D~\cite{lin2024sam} detects the objects with a similar strategy as CNOS~\cite{nguyen2023cnos} but computes a weighted score including semantics, appearance, and geometry to match the query object template with the segments extracted from SAM~\cite{kirillov2023segment}.

\subsubsection{Results.} 
\begin{figure}[t]
    \centering
        \centering
    \includegraphics[width=\textwidth]
    {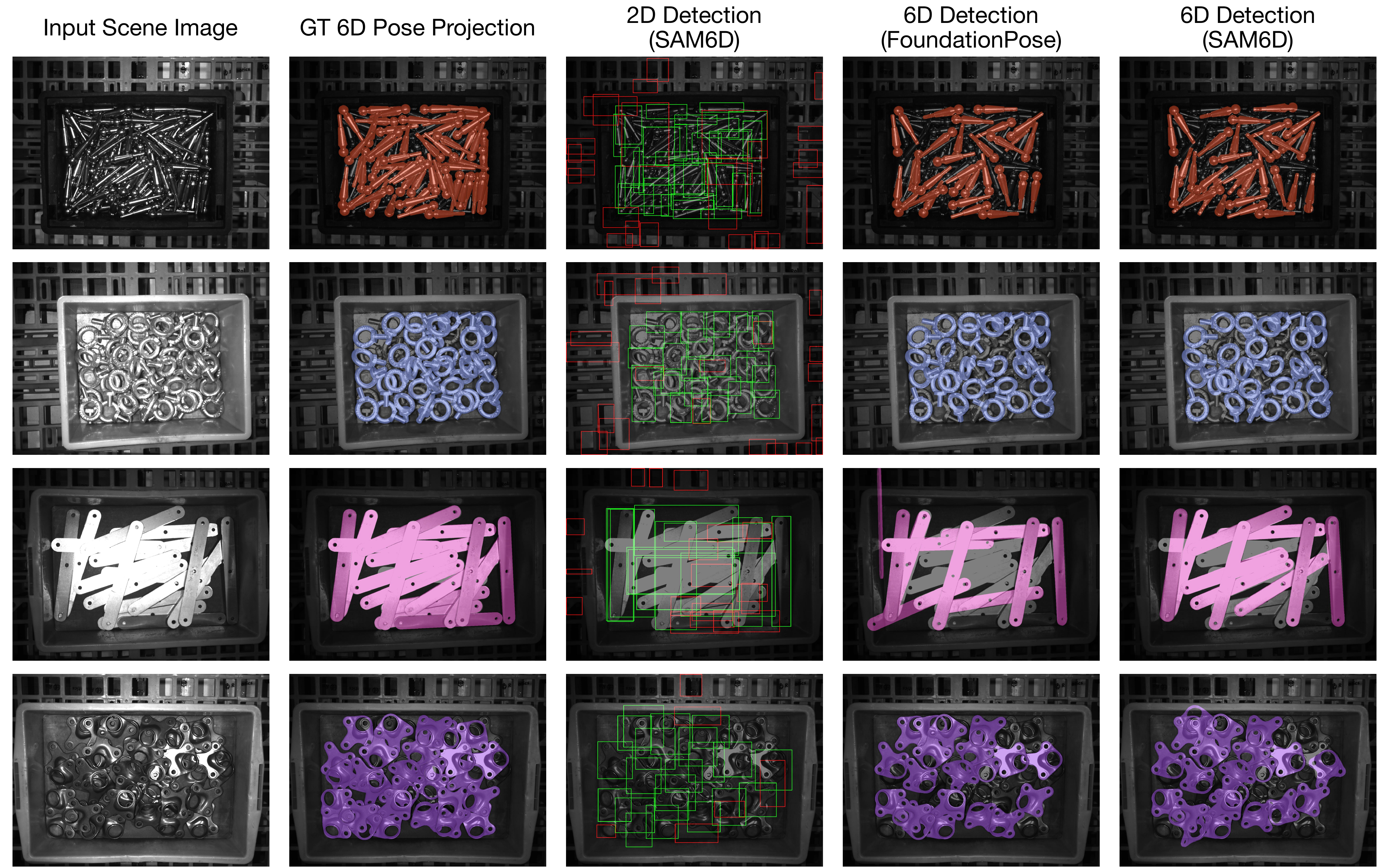} 
    \caption{ \small Example qualitative results of SOTA methods for unseen object 2D and 6D detection tasks on XYZ-IBD dataset. In the detection results, green boxes indicate correct object IDs, while red boxes indicate mismatches with the ground truth ID. } 
    \label{fig:visualization}
\end{figure}

As shown in Table~\ref{tab:pose_baselines}, for the 2D detection task, YOLOX~\cite{ge2021yolox} is trained on the training split of each dataset and tests on the test split for the same objects (seen objects), achieving comparable results across different datasets under this overfitting setup. However, for the unseen object detection methods, we directly deploy the pretrained model to inference on our dataset. CNOS and SAM6D show a clear drop compared to other BOP datasets. Specifically, SAM6D degrades by more than 33\% of the AP on average. These results highlight the increased difficulty of our dataset for 2D detection, due to heavy occlusion, repeated object instances, and strong surface reflections. We show a qualitative comparison in Figure~\ref{fig:visualization}.

\subsection{Evaluation of Object 6D Detection}
\subsubsection{Seen Object 6D Detection.} SurfEmb~\cite{haugaard2022surfemb} learns per-object dense 2D–3D correspondence distributions over object surfaces using contrastive learning in an unsupervised fashion. It achieves strong performance on BOP and handles visual ambiguities effectively.
GDRNet~\cite{wang2021gdr} is a recent state-of-the-art framework that processes zoomed-in RoIs from RGB images to predict intermediate geometric features: dense 2D-3D correspondences, surface region attention maps, and visible object masks. These features guide a Patch-PnP module to directly regress the 6D pose in a differentiable manner. 

\subsubsection{Unseen Object 6D Detection.} FoundationPose~\cite{wen2024foundationpose} supports both model-based and model-free settings using neural implicit representations for view synthesis. Trained on large-scale synthetic data with transformer-based coarse-to-fine design, it generalizes well and outperforms prior methods across benchmarks. MatchU~\cite{huang2024matchu} extracts texture and geometric features from the input RGB-D and CAD models to fuse the descriptors for 3D registration. Similarly,  
SAM-6D~\cite{lin2024sam} uses the Segment Anything Model for segmentation and applies ViT~\cite{vit} and GeoTransformer~\cite{qin2023geotransformer} to extract features from RGB-D input and CAD models. Trained on a large-scale synthetic dataset, it achieves strong performance in model-based 6D pose estimation.

\subsubsection{Results.} Unseen object methods assume the availability of object segmentation or detection as a prior for pose estimation. Accordingly, we use segmentation masks produced by SAM-6D for fair comparison among these methods, while seen object methods utilize detection results from YOLOX. As shown in Table~\ref{tab:pose_baselines}, all methods struggle on our dataset. Specifically, both GDRNet and SurfEmb, representing seen object methods, fail to predict accurate poses, despite being trained on synthetic data. In contrast, unseen object methods demonstrate relatively better performance, with SAM6D achieving state-of-the-art results. Compared to existing household datasets, ~\cite{van2025bop,sundermeyer2023bop}, our dataset introduces greater challenges for pose estimation due to the complexity of object materials, geometric variations, and severe scene clutter. Figure~\ref{fig:visualization} shows a qualitative comparison of the baseline results on XYZ-IBD benchmark.

\section{Conclusion}
\label{sec:conclusion_and_limitation}

We introduce XYZ-IBD, a high-precision benchmark for industrial bin-picking designed to capture the complexities of real-world manufacturing, including stochastic scene configurations, multi-instance ambiguity, and specular reflectivity. The benchmark comprises a diverse set of industrial components captured under authentic factory conditions using three distinct sensors, yielding 273k annotated real-world samples. This is augmented by a 45k-frame synthetic dataset that simulates realistic bin-picking environments. Utilizing a multi-stage, semi-automatic annotation protocol, we provide high-fidelity 6D pose labels. We rigorously quantify annotation error through simulations that model sensor and calibration noise, achieving a mean pose error below 1 mm. We believe XYZ-IBD brings real-world industrial vision problems to the academic community and helps bridge the gap between academic research and practical application. Potential future work lies to expand the scale of object categories and working conditions, covering more typical but challenging vision tasks for the evaluation of recent foundational industrial solutions.

\section{Acknowledgement}
\label{sec:aknowledgement}
We thank XYZ Robotics for providing the industrial parts and environments used for dataset collection. We are especially grateful to Jizhong Liang for his valuable efforts in annotating the dataset during his internship at XYZ Robotics. We also thank Tomáš Hodaň for his help in integrating the XYZ-IBD dataset into the BOP benchmark and Médéric Fourmy for assisting with the evaluation system integration.

\bibliographystyle{splncs04}
\bibliography{main}
\end{document}